\documentclass[10pt,twocolumn,letterpaper]{article}

\usepackage{iccv}
\usepackage{times}
\usepackage{epsfig}
\usepackage{graphicx}

\usepackage{amssymb}
\usepackage{url}
\usepackage{graphicx}
\usepackage{booktabs}
\usepackage{float}
\usepackage{multirow}
\usepackage{makecell}

\usepackage[accsupp]{axessibility}  

\usepackage[breaklinks,colorlinks]{hyperref}
\usepackage[capitalize]{cleveref}
\iccvfinalcopy 

\crefname{section}{Sec.}{Secs.}
\Crefname{section}{Section}{Sections}
\Crefname{table}{Table}{Tables}
\crefname{table}{Tab.}{Tabs.}

\usepackage[dvipsnames]{xcolor}
\usepackage{hyperref}
\usepackage{amsmath}
\hypersetup{
	final,
	breaklinks=true,
	colorlinks,
	linkcolor={Mahogany},
	citecolor={YellowGreen},
	urlcolor={CadetBlue}
}

\ificcvfinal\fi

\begin{document}

\title{CL-MVSNet: Unsupervised Multi-view Stereo with \\
Dual-level Contrastive Learning}

\author{Kaiqiang Xiong \textsuperscript{1} \quad Rui Peng \textsuperscript{1} \quad Zhe Zhang \textsuperscript{1} \quad Tianxing Feng \textsuperscript{1} \\
Jianbo Jiao \textsuperscript{4} \quad Feng Gao \textsuperscript{5} \quad Ronggang Wang \textsuperscript{1,2,3}\\
\textsuperscript{1}School of Electronic and Computer Engineering, Peking University \\
\textsuperscript{2}Peng Cheng Laboratory \quad \textsuperscript{3}Migu Culture Technology Co., Ltd \\ \textsuperscript{4}School of Computer Science, University of Birmingham \quad \textsuperscript{5}School of Arts, Peking University \\
{\tt\small xiongkaiqiang@stu.pku.edu.cn \quad rgwang@pkusz.edu.cn} \\
{ \url{https://KaiqiangXiong.github.io/CL-MVSNet/}}}


\maketitle
\ificcvfinal\thispagestyle{empty}\fi

\begin{abstract}
Unsupervised Multi-View Stereo (MVS) methods have achieved promising progress recently.
However, previous methods primarily depend on the  photometric consistency assumption, which may suffer from two limitations: indistinguishable regions and view-dependent effects, \eg, low-textured areas and reflections. 
To address these issues, in this paper, we propose a new dual-level contrastive learning approach, named \textbf{CL-MVSNet}. 
Specifically, our model integrates two contrastive branches into an unsupervised MVS framework to construct additional supervisory signals. 
On the one hand, we present an image-level contrastive branch to guide the model to acquire more context awareness, thus leading to more complete depth estimation in indistinguishable regions. 
On the other hand, we exploit a scene-level contrastive branch to boost the representation ability, improving robustness to view-dependent effects. 
Moreover, to recover more accurate 3D geometry, we introduce an $\mathcal{L}0.5$ photometric consistency loss, which encourages the model to focus more on accurate points while mitigating the gradient penalty of undesirable ones.
Extensive experiments on DTU and Tanks\&Temples benchmarks demonstrate that our approach achieves state-of-the-art performance among all end-to-end unsupervised MVS frameworks and outperforms its supervised counterpart by a considerable margin without fine-tuning. 
\end{abstract}

\section{Introduction}
\label{sec:introduction}

\begin{figure}[t]
  \begin{center}
     \includegraphics[trim={0cm 0cm 0cm 0cm},clip,width=1.0\linewidth]{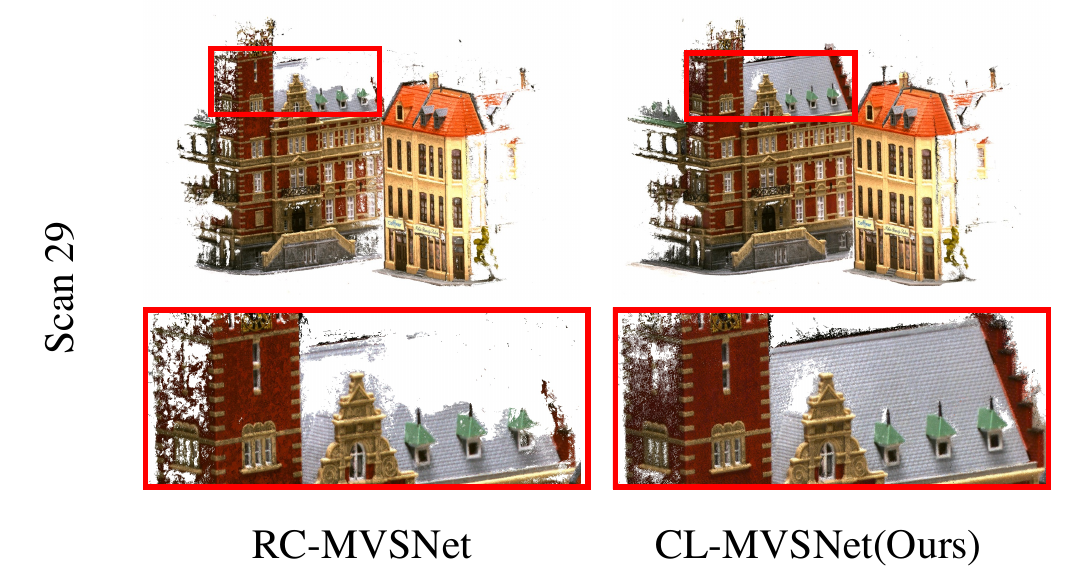}
  \end{center}
  \vspace{-0.3cm}
  \caption{\textbf{Qualitative comparison of reconstruction quality with the SOTA 
 method \cite{chang2022rc} on scan29 of DTU \cite{aanaes2016large}.} Our method performs better on repetitive patterns.
 }
  \label{fig:head}
  \vspace{-0.3cm}
\end{figure}

Multi-View Stereo (MVS) is a critical task in various applications, including robotics, self-driving, and VR/AR. 
The goal of MVS is to estimate a dense 3D reconstruction from multiple images captured from different views.
Traditionally, it has been approached by computing dense correspondences between images, often based on hand-crafted similarity metrics and engineered regularizations \cite{barnes2009patchmatch, seitz2006comparison}.
Recently, a surge of learning-based  methods \cite{Yao_2018_ECCV,yao2019recurrent,yang2020cost,cheng2020deep} have been developed to advance the effectiveness of MVS, showing promising results in MVS benchmarks \cite{aanaes2016large,knapitsch2017tanks}.
However, most of them are supervised methods \cite{mi2022generalized,ding2022transmvsnet,bae2022multi}. These methods heavily rely on large-scale ground-truth 3D training data, which are expensive to acquire.

To tackle this problem, unsupervised MVS methods \cite{khot2019learning,chang2022rc,xu2021self} have attempted to train MVS networks without annotations. 
Existing methods mainly depend on the hypothesis of photometric consistency, which states that the appearance of a point in 3D space is invariant across different views.
However,  this hypothesis may be ineffective owing to indistinguishable regions and view-dependent effects, \eg, low texture, repetitive patterns, and reflections.
Recently, the state-of-the-art (SOTA) method RC-MVSNet \cite{chang2022rc} adopts a rendering consistency network to address the ambiguity caused by view-dependent photometric effects and occlusions.
While achieving promising results, this approach may suffer from significant performance degradation in indistinguishable regions, as shown in \cref{fig:head}.
To address the absence of valid supervisory signals, we propose to leverage contrastive learning to boost the robustness and generalizability of unsupervised MVS in various scenarios.

Contrastive learning is a widely used paradigm in unsupervised learning to construct additional supervisory signals.
With the success of contrastive learning in images, \cite{xu2021self,zhangelasticmvs} have extended its application to unsupervised MVS with color fluctuation augmentations or part segmentations. 
However, there are still many remaining unresolved issues, \eg, occlusions and low-textured surfaces.
And the potential of incorporating contrastive learning into MVS remains largely under-explored.
Moreover, recent studies \cite{afham2022crosspoint,zhu2021improving} showed that {\em hard positive samples are of benefit to boost the contrastive learning}.
Inspired by this, we propose a dual-level contrastive learning approach, named \textbf{CL-MVSNet}, where image-level and scene-level contrastive learning  are integrated into an unsupervised MVS framework.

To resolve ambiguity from indistinguishable regions, we introduce an image-level contrastive learning strategy to encourage the model to be more context-aware.
Specifically, for an image-level contrastive sample, all pixels in the source images are masked with independent and identically distributed Bernoulli probability, simulating the case that local photometric consistency fails.
Following that, we maximize the similarity between the depth estimations of the regular sample and the image-level contrastive sample.
The intuition is that the augmented images contain the same context information as the original ones, which can also be utilized to estimate complete depth maps as hard positive samples.
In this way, the network is encouraged to exploit more contextual information instead of relying only on photometric consistency over small regions.

In addition, we propose a scene-level contrastive learning branch to alleviate the view-dependent photometric effects.
Due to severe occlusions, reflections, and illumination changes, source images with few overlaps are often infeasible to use in unsupervised MVS.
However, from the perspective of contrastive learning, a scene-level contrastive sample containing randomly selected source images can be considered a natural hard positive sample. 
These hard samples are expected to produce the same 3D representation as
the regular samples, as they are captured from identical scenes.
Therefore, we enforce contrastive consistency between the scene-level contrastive branch and the regular branch, encouraging the model to learn more powerful 3D cost volume regularization for robust depth estimation.


Furthermore,  we propose an $\mathcal{L}0.5$ photometric consistency loss to support the training of the CL-MVSNet.
In an MVS system, after depth estimation, most points with undesirable depth predictions will be filtered out before depth fusion.
Besides, these points often locate in occluded areas or useless backgrounds where photometric consistency enforcement may mislead the model.
To this end, instead of using vanilla photometric consistency loss based on $\mathcal{L}1$ or $\mathcal{L}2$ norm, we propose to use the $\mathcal{L}0.5$ norm, which has larger gradients with regard to smaller errors.
In this way, the model increases the penalty of accurate points, resulting in more accurate survival points.

In conclusion, our main contributions are: 

\begin{itemize}
  \item  We present an image-level contrastive consistency loss, which encourages the model to be more context-aware and recover more complete reconstruction in indistinguishable regions.
  \item  We propose a scene-level contrastive consistency loss, which boosts the representation ability to promote robustness to view-dependent effects.
  \item  We propose an $\mathcal{L}0.5$ photometric consistency loss to further advance the contrastive learning framework, which enables the model to focus on accurate points, resulting in more accurate reconstruction.
  \item Experiments on DTU \cite{aanaes2016large} and Tanks\&Temples \cite{knapitsch2017tanks} benchmarks show that our method outperforms state-of-the-art end-to-end unsupervised models and surpasses its supervised counterpart.
\end{itemize}

\begin{figure*}[t]
  \vspace{-0.5cm}
  \begin{center}
     \includegraphics[width=1.0\linewidth]{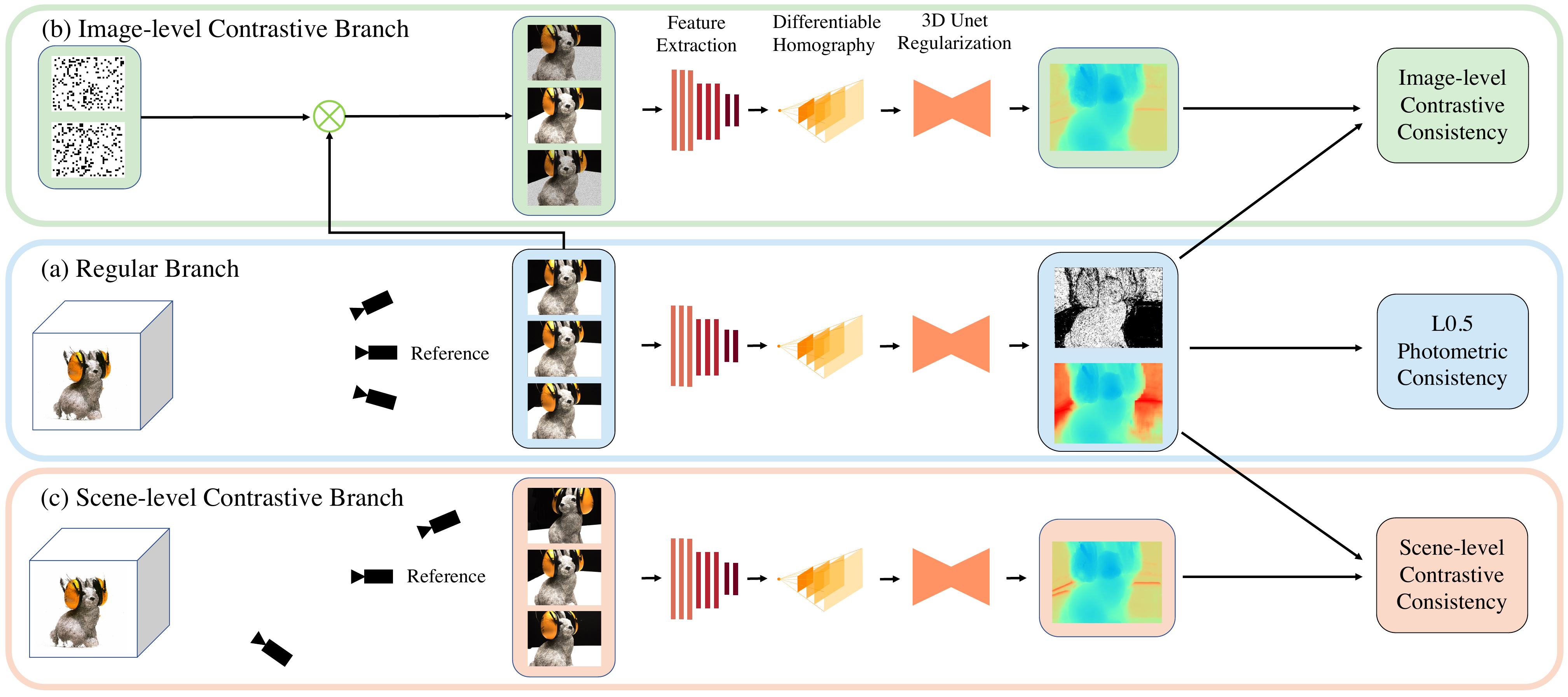}
  \end{center}
  \vspace{-0.3cm}
  \caption{\textbf{The framework of CL-MVSNet.} The framework consists of: (a) a Regular Branch with a regular sample similar to CasMVSNet \cite{gu2020cascade}, (b) an Image-level Contrastive Branch with the image-level contrastive sample
  , (c) a Scene-level Contrastive Branch with the scene-level contrastive sample. 
 To pull positive pairs close, we enforce contrastive consistency between the regular branch and two contrastive branches, with the confidence mask estimated from the regular branch. 
 Moreover, our proposed $\mathcal{L}0.5$ photometric consistency is enforced between the reconstructed images and the input reference image on the regular branch for more accurate reconstruction.
 }
  \label{fig:framework}
  \vspace{-0.4cm}
\end{figure*}

\section{Related Work}
\label{sec:related_work}
\noindent \textbf{Supervised MVS.} With the development of deep learning technique and large-scale 3D datasets, supervised MVS has achieved significant progresses in recent years \cite{Yao_2018_ECCV,yao2019recurrent,luo2019p,xu2019multi,zhang2020visibility,yang2020cost,xu2020learning,wang2021patchmatchnet,peng2022rethinking,wang2022mvster,wang2022mvsnet,yang2022non,ma2022multiview,xi2022raymvsnet,ma2021epp}. 
MVSNet \cite{Yao_2018_ECCV} proposes a popular MVS pipeline, which can be summarized as four steps: feature extraction, cost aggregation, cost volume regularization, and depth regression.
Recent works make efforts to relieve the huge memory and computation cost, by introducing a multi-stage architecture \cite{cheng2020deep,gu2020cascade,yang2020cost,mi2022generalized}, RNN \cite{yao2019recurrent} and some other methods \cite{wang2021patchmatchnet,wang2022mvster}.
However, all the above approaches depend on labeled training data, which are expensive to obtain in practice.

\

\noindent \textbf{End-to-end Unsupervised and Multi-stage Self-supervised MVS.}
Following \cite{chang2022rc}, we categorize current multi-view stereo methods trained without any annotations into two dominant groups: the end-to-end unsupervised MVS and the multi-stage self-supervised MVS.
The end-to-end unsupervised MVS methods \cite{khot2019learning,dai2019mvs2,huang2021m3vsnet,xu2021self,chang2022rc,li2022ds} primarily rely on the hypothesis of photometric consistency.
Concretely, Unsup-MVSNet proposes the first end-to-end unsupervised MVS framework. 
It enforces photometric consistency between the reconstructed images and the reference image. 
In addition, it utilizes structured similarity loss and depth smoothness loss to further improve the performance.
However, photometric consistency is ineffective in many challenging areas. 
Thus, recent researchers have attempted to incorporate pretext tasks into training to find reliable supervisory signals, \eg, normal-depth consistency \cite{huang2021m3vsnet}, semantic consistency \cite{xu2021self}, and rendering consistency \cite{chang2022rc}.
However, there is still a performance gap between unsupervised methods and supervised SOTA methods.

While multi-stage self-supervised methods \cite{yang2021self,xu2021digging,ding2022kd,qiu2022self} aim to obtain reliable pseudo-labels by filtering and processing inferred depth maps, which are used to supervise model training in subsequent stages. 
However, these methods cannot be trained in an end-to-end manner since they command complex pre-processing and fine-tuning. 
Additionally, the generated pseudo-labels require extra storage space, and iterative self-training can take significant time.

\

\noindent \textbf{Contrastive Learning.}
Contrastive learning \cite{chen2020simple,grill2020bootstrap,he2020momentum,misra2020self} is a popular paradigm in unsupervised learning that encourages the model to be invariant to multiple transformations of a single sample.
Specifically, contrastive learning pulls positive sample pairs close while pushing negative sample pairs away. 
This approach has achieved remarkable success in narrowing the performance gap between unsupervised and supervised models.
Recently, hard positive samples have been confirmed to be beneficial for boosting contrastive learning \cite{zhu2021improving,afham2022crosspoint}. 
For this purpose, some studies have explored pixel- \cite{xie2021propagate}, image- \cite{he2020momentum}, and object-level \cite{li2022contextual} contrastive learning. 
Most image-level contrastive learning methods rely on well-designed augmentation procedures.
However, pixel- and object-level supervision provides a way to directly define positive and negative samples.
In this work, we propose to leverage a dual-level contrastive learning strategy to boost the unsupervised MVS, which includes image-level and scene-level contrastive learning.

\section{Method}
\label{sec:method}

In this section, the main contributions of CL-MVSNet  will be elaborated.
We firstly depict the unsupervised backbone (in \cref{sec:backbone}), then we describe the proposed image-level contrastive consistency (in \cref{sec:icc}),  the scene-level contrastive consistency (in \cref{sec:scc}), and the $\mathcal{L}0.5$ photometric consistency (in \cref{sec:l0.5pc}).
Finally, we introduce the overall loss function during training (in \cref{sec:optimization}).
An overview of our architecture is shown in \cref{fig:framework}.
Note that CL-MVSNet is a general framework suitable for arbitrary learning-based MVS.
And we take the representative CasMVSNet \cite{gu2020cascade} as our backbone in this work. 

\subsection{Unsupervised Multi-view Stereo}
\label{sec:backbone}

To begin with, we adopt the same view-selection strategy as CasMVSNet \cite{gu2020cascade} to construct a regular sample.
For a given regular sample comprised of $1$ reference image $I_1$, $N-1$ source images $\{I_i\}_{i=2}^{N}$ and their camera parameters $\{K_i, T_i\}_{i=1}^{N}$, our goal is to estimate the corresponding depth map with the backbone network.
Specifically, the backbone consists of four steps: feature extraction, cost volume construction, cost volume regularization, and depth regression.

During feature extraction, images $\{I_i\}_{i=1}^{N}$  are fed into a shared Feature Pyramid Network \cite{lin2017feature} to generate 2D pixel-wise features $\{F_i\}_{i=1}^{N}$ at three stages with incremental resolutions.
At the coarsest stage, with initial depth hypothesis $\{d_{min}, \ldots,d_{max}\}$, 3D feature volumes $\{V_i\}_{i=1}^{N}$ are built from features $\{F_i\}_{i=1}^{N}$ via differentiable homography. 

As for cost volume aggregation, we adopt group-wise correlation metric following \cite{wang2021patchmatchnet,xu2020learning,zhang2020visibility,mi2022generalized}. 
We divide the $N$ feature volumes $\{V_i\}_{i=1}^{N}$  of $N_C$ channels into $N_G$ groups, then construct the raw cost volume $C$ as below:

\begin{equation}
  \mathbf{C} = \frac{1}{(N-1)N_C/N_G}\sum_{i=2}^{N-1}\langle V_1^g,V_i^g \rangle_{g=1}^{N_G}.
\label{eq:cost_volume}
\end{equation}

Then the raw cost volume $\mathbf{C}$ undergoes a regular  3D U-Net  and a softmax, resulting in a probability volume $P_v$.
Finally, the depth map $D$ is obtained by weighted sum:
\begin{gather}
D=\sum \limits _{d=d_{min}} ^{d_{max}}d \times P_v(d).
\end{gather}

In a coarse-to-fine fashion, the coarse depth map $D^1$ is used to generate the depth hypothesis of the next stage. 
With features $\{F_i^2, F_i^3\}_{i=1}^{N}$ at the larger resolution, finer depth maps $\{D^2, D^3\}$ will be  estimated iteratively.

\

\noindent \textbf{Depth estimation with Confidence Mask.} During the depth estimation, there is a byproduct probability volume $P_v$, measuring the pixel-wise confidence of the depth hypothesis.
Then a probability map $P_m$ can be acquired by taking the probability sum over the four nearest depth hypotheses with regard to depth estimation \cite{Yao_2018_ECCV}.
We generate a binary confidence mask $M_c$ that indicates whether the model is confident about the pixel-wise depth estimation:
\begin{gather}
M_c = P_m > \gamma,
\end{gather}
where the $\gamma$ is set to 0.95 in our implementation.
The depth estimation $D_R$ from the regular branch and the confidence mask $M_c$ will be used to regularize the outputs of the two proposed contrastive branches (in \cref{sec:icc} and \cref{sec:scc}).

\

\begin{figure}[t]
  \begin{center}
     \includegraphics[trim={0cm 3cm 0cm 1cm},clip,width=1.0\linewidth]{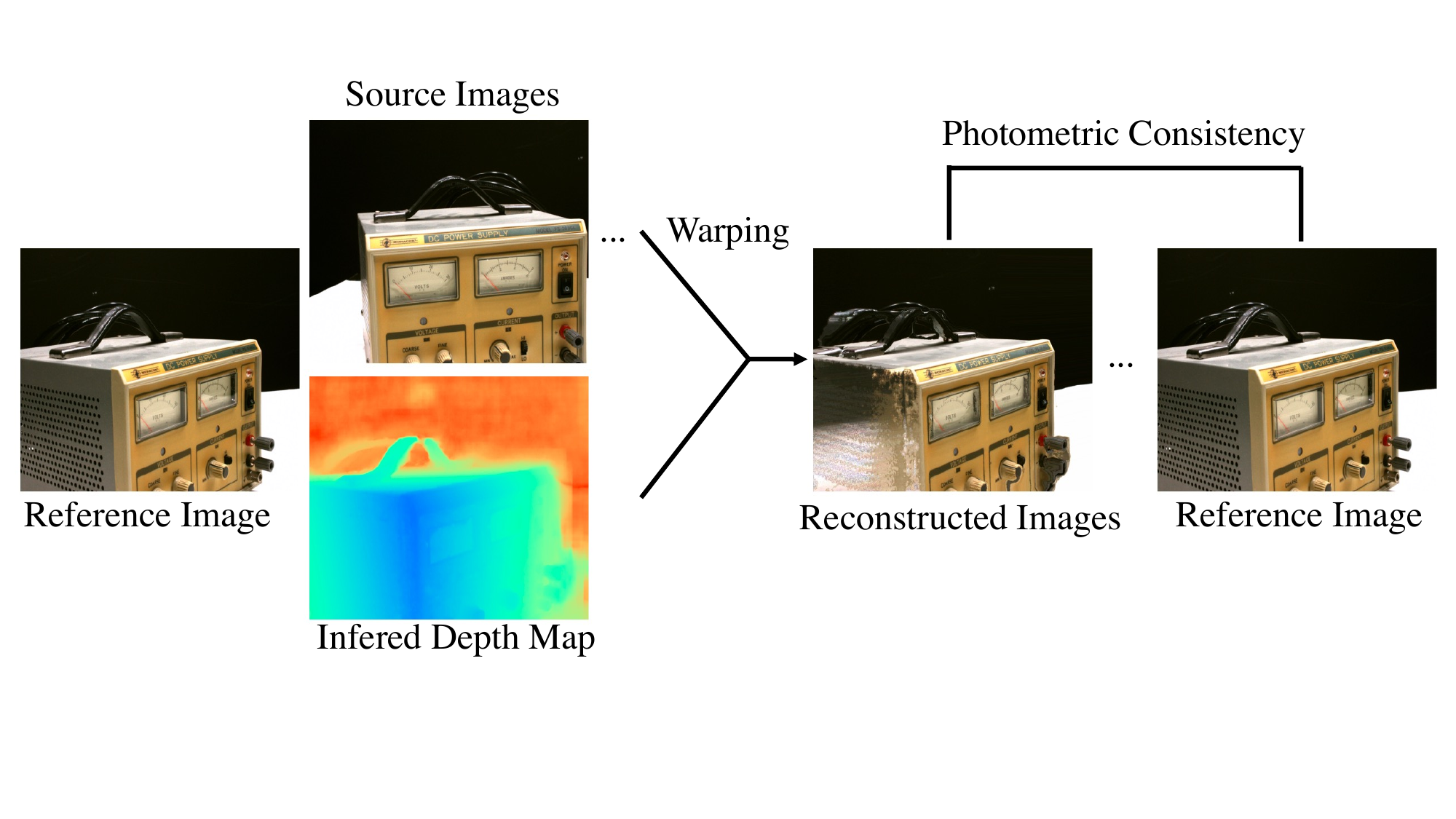}
  \end{center}
  \vspace{-0.2cm}
  \caption{\textbf{Photometric consistency.} The source images are warped to reconstruct the reference image with the inferred depth map on the reference view. Then consistency is enforced between the reconstructed images and the reference image.
  }
  \label{fig:pc}
  \vspace{-0.1cm}
\end{figure}

\noindent \textbf{Vanilla Photometric Consistency Loss.} Previous unsupervised MVS methods \cite{khot2019learning,qiu2022self,yang2021self,xu2021digging,ding2022kd,chang2022rc,xu2021self,dai2019mvs2,huang2021m3vsnet} use a photometric consistency loss to train an unsupervised MVS network without any ground truth depth, as shown in \cref{fig:pc}.

Specifically, given a reference image $I_1$, a source image $I_i$, associated intrinsic $K$, relative transformation $T$, and an inferred depth map $D_R$, the source image $I_i$ is warped to reconstruct reference image $\hat{I}_i$.
For a specific pixel $p$ in reconstructed reference image $\hat{I}_i$, its coordinate $p'$ in source image $I_i$ can be calculated via inverse warping:  
\begin{gather}
p'=KT(D_{R}(p) \cdot K^{-1}p).
\end{gather}

Then the reconstructed image $\hat{I}_i$ can be obtained via differentiable bilinear sampling $ \hat{I}_i(p)=I_i(p')$.

Along with the inverse warping process, a binary valid mask $M_i$ is generated, indicating valid pixels in the reconstructed image $\hat{I}_i$. 
In previous unsupervised MVS methods, all source images $\{I_i\}_{i=2}^{N}$ are warped to the reference view according to the inferred reference depth $D_R$, then the vanilla photometric consistency loss can be computed as:
\begin{equation}
L_{PC}=\sum \limits _{i=2}^{N} \frac{\Vert (\hat{I}_i-I_1)\odot M_i \Vert_{\theta} \\ + \Vert(\nabla \hat{I}_i-\nabla I_1) \odot M_i \Vert_{\theta} }{\Vert M_i \Vert_1},
\label{eq:vanilla_photometric_loss}
\end{equation}
where $\nabla$ refers to the gradient operator, $\odot$ is element-wise product, $\theta$=1 or 2, which denotes the $\mathcal{L}1$ or $\mathcal{L}2$ norm.

\subsection{Image-level Contrastive Consistency}
\label{sec:icc}
As mentioned before, the local photometric consistency fails to offer valid supervisory signals in indistinguishable regions, \eg, areas with low texture or repetitive patterns.
To overcome this problem, \cite{dong2022patchmvsnet} explicitly introduces a patch-wise photometric consistency loss to enhance the matching robustness.
However, the model may suffer notable performance degradation for inappropriate handcrafted patch size, \eg, the large patch may lead to the loss of accuracy in rich-textured areas.
In this work, we propose an alternative to implicitly encourage the model to be context-aware by introducing an image-level contrastive branch.
The image-level contrastive sample and image-level contrastive consistency loss will be elaborated on next.

\begin{figure}[t]
  \begin{center}
     \includegraphics[trim={0cm 0.2cm 0cm 0cm},clip,width=0.8\linewidth]{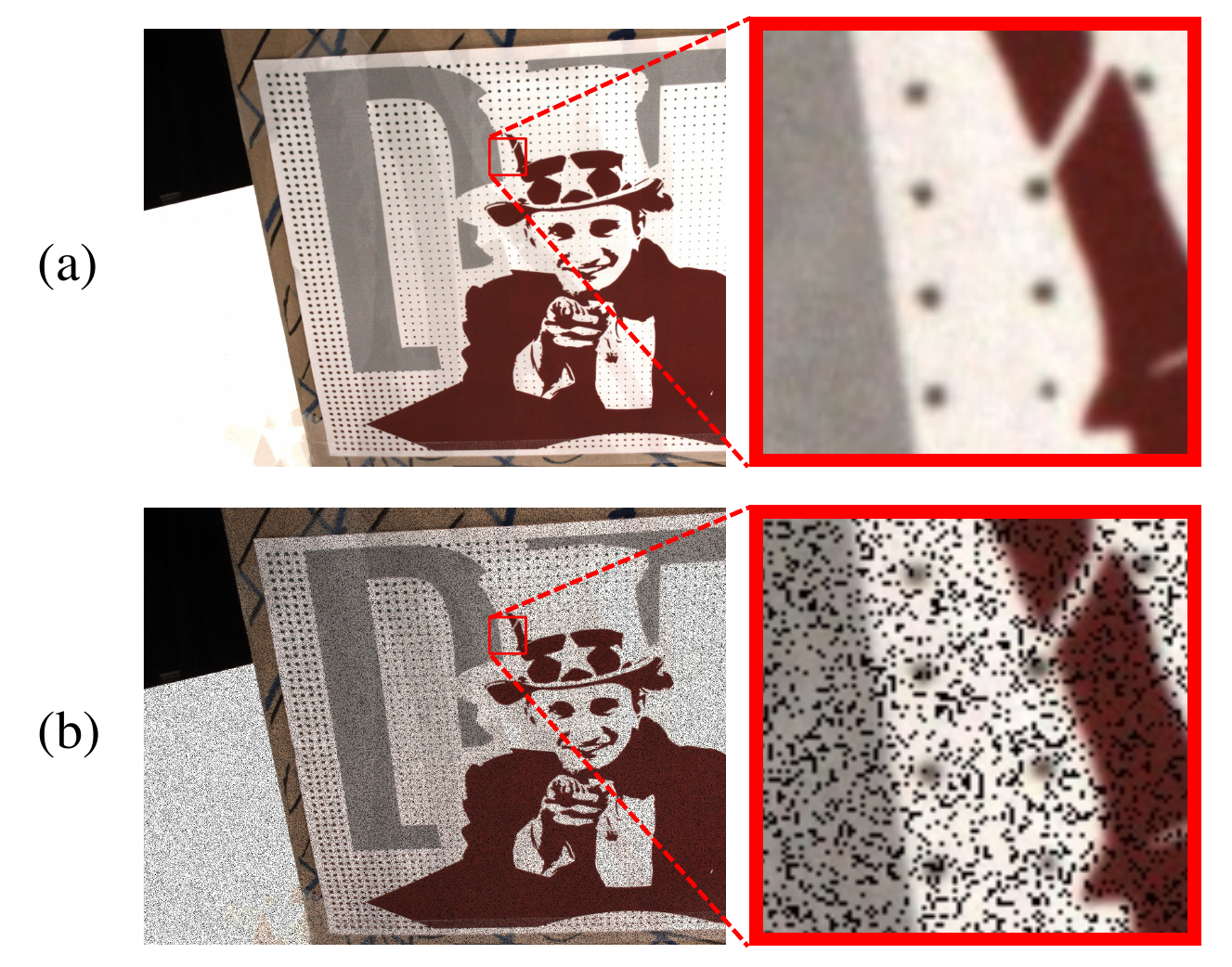}
  \end{center}
  \vspace{-0.2cm}
  \caption{\textbf{Image-level contrastive sample.} (a) a source image of the regular sample. (b) a source image of the image-level contrastive sample. 
 A Bernoulli-distributed binary mask is used to simulate the failure case of local photometric consistency in (b).}
  \label{fig:ic}
  \vspace{-0.3cm}
\end{figure}

\

\noindent \textbf{Image-level Contrastive Sample.}
The image-level contrastive sample is generated by applying transformation  on a  given regular sample $\{ I_i\}_{i=1}^N$ artificially.
For a source image $I_i$ with the size of $H\times W \times C$, we set an occlusion rate $\alpha$ to construct a binary pixel-wise mask with the size of $H\times W$.
Concretely, any element in the mask will draw a value of 1 according to the given occlusion rate $\alpha$:
\begin{gather}
M_{o(i,j)} \thicksim B(\alpha),
\end{gather}
where $B$ denotes the Bernoulli distribution, the occlusion mask is denoted as $M_o$.
Then $M_o$ is used to mask the image $I_i$ as shown in \cref{fig:ic}.
Simply processing all source images in this way, an image-level contrastive sample $\{ I_1, I'_2, I'_3, \ldots , I'_N\}$ is generated.
Then the hard positive sample will be fed to the net to obtain the depth estimation $D_{IC}$.
To ensure training stability, we employ a curriculum learning strategy to gradually increase the occlusion rate $\alpha$, which grows from 0 to 0.1 in our implementation.

\

\noindent \textbf{Image-level Contrastive Consistency Loss.}
To enforce consistency between the regular branch and the image-level contrastive branch, we compute the image-level contrastive consistency loss $L_{ICC}$ as:
\begin{gather}
L_{ICC}=\frac{ \Vert (D_R - D_{IC}) \odot M_c \Vert_1}{ \Vert {M_c} \Vert_1 },
\end{gather}
where $D_R$ and $D_{IC}$ denote the inferred depth of the regular branch and the image-level contrastive branch respectively.

\subsection{Scene-level Contrastive Consistency}
\label{sec:scc}

\begin{figure}[t]
  \begin{center}
     \includegraphics[trim={2cm 2cm 2cm 2cm},clip,width=1.0\linewidth]{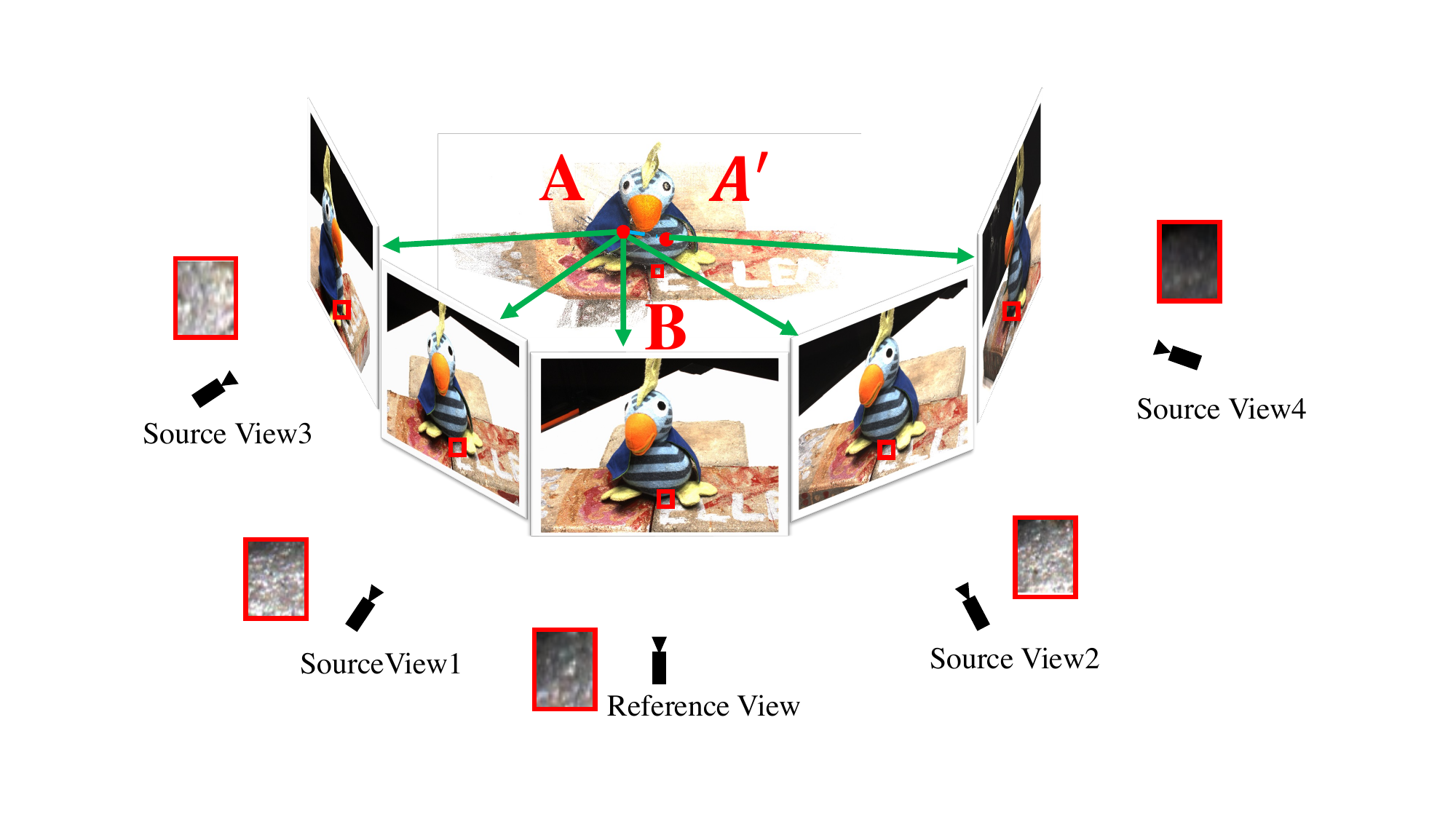}
  \end{center}
  \vspace{-0.3cm}
  \caption{\textbf{View-dependent effects and occlusions.} From source view4, point $A$ is occluded and only point $A'$ is visible along the line of sight. 
  Besides, The appearance of the identical region $B$ differs in different views due to variations in illumination, camera exposure, and reflections.}
  \label{fig:example}
  \vspace{-0.4cm}
\end{figure}

To the best of our knowledge, all unsupervised MVS methods including our regular sample apply photometric and geometric priors for view selection following MVSNet \cite{Yao_2018_ECCV}.
Specifically, images with more overlaps with the reference image will get higher scores, and the $N-1$ images with the highest scores will be selected as source images in the regular sample.
Note that some supervised methods \cite{barnes2009patchmatch, wang2022itermvs} randomly select source views to improve the robustness.
Due to severe view-dependent effects and occlusions where local photometric consistency fails (in \cref{fig:example}), directly using this strategy in the regular branch of unsupervised MVS will lead to worse performance.
However, from a contrastive learning perspective, these images can be used to construct hard positive samples.
Below, we will introduce how to generate the scene-level contrastive sample and enforce the scene-level contrastive consistency in detail.

\

\noindent \textbf{Scene-level Contrastive Sample.} For a given reference image $I_1$, the scene-level contrastive sample can be constructed by combining the reference image $I_1$with $N-1$ randomly selected image $\{I''_i\}_{i=2}^N$ of the same scene.
Afterward, a depth map $D_{SC}$ will be inferred for the contrastive sample.
It is worth noting that the cost volume built with the scene-level contrastive sample represents the identical 3D scene as the regular ones.
Thus, the network is supposed to gain the same depth estimation from the scene-level contrastive sample as the regular ones.

\

\noindent \textbf{Scene-level Contrastive Consistency Loss.} Then a scene-level contrastive consistency loss $L_{SCC}$ will be applied on the scene-level contrastive branch. 
The loss is used to pull the scene-level contrastive sample closer to the regular sample, improving the robustness to view-dependent effects:
\begin{gather}
L_{SCC}=\frac{ \Vert (D_R - D_{SC}) \odot M_c \Vert_1}{ \Vert {M_c} \Vert_1 },
\end{gather}
where $D_R$ and $D_{SC}$ refer to the depth map from the regular branch and the scene-level contrastive branch respectively.

\subsection{\texorpdfstring{$\mathcal{L}0.5$ Photometric Consistency}{L0.5 Photometric Consistency}}
\label{sec:l0.5pc}

\begin{figure}[t]
  \vspace{-0.5cm}
  \begin{center}
     \includegraphics[trim={0cm 0cm 0cm 0cm},clip,width=1.0\linewidth]{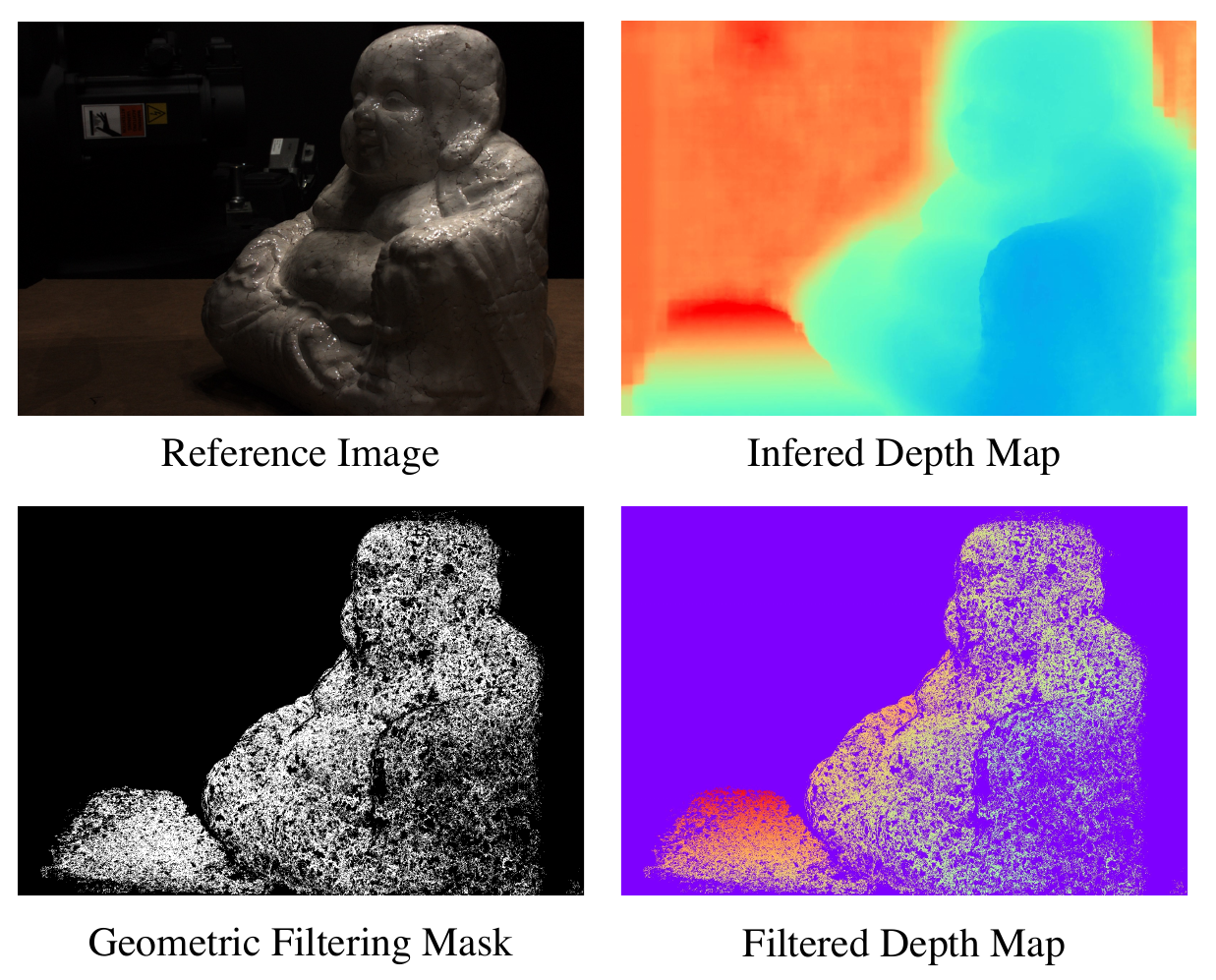}
  \end{center}
  \vspace{-0.3cm}
  \caption{\textbf{Depth filtering.} Most points with inaccurate depth will be filtered out before depth fusion. It can be observed that these points usually locate in occluded regions or useless backgrounds, where photometric consistency fails.}
  \label{fig:filtering}
  \vspace{-0.4cm}
\end{figure}

In the generic pipeline of MVS, a depth filtering process is applied before depth fusion, in order to mask out most undesirable points, as shown in \cref{fig:filtering}. 
For clarity, we divide the points in the inferred depth map into three categories:
\begin{itemize}
  \vspace{-0.1cm}
\setlength{\itemsep}{2pt}
\setlength{\parsep}{0pt}
\setlength{\parskip}{0pt}
  \item  {\em accurate point}: the pixel with accurate prediction, which will survive the depth filtering and contribute to the final reconstruction. 
  \item  {\em ordinary point}: the pixel close to the accurate point, which will be filtered out.
  \item  {\em terrible point}: the pixel with pretty erroneous prediction, most of which locate in useless backgrounds or occluded areas and will be filtered out. 
  Moreover, the photometric consistency enforced on these points may mislead the model.
    \vspace{-0.1cm}
\end{itemize}

In the MVS system, an ideal depth map is supposed to contain more {\em accurate points}.
Moreover, a loss function that imposes more constraints on {\em accurate points} will be desirable due to this observation.
We perform an analysis of the $\mathcal{L}1$-norm and $\mathcal{L}2$-norm, which are broadly used in vanilla photometric consistency loss (\cref{eq:vanilla_photometric_loss}).
We first write their expressions and gradient formulas:
\begin{equation}
    \left\{
    \begin{aligned}
        l_1(e) & = \Vert e \Vert_1  = (\sum \limits _{i=1}^{n} e_{i}),  \frac{\partial l_1(e) }{\partial e_i}  =  1, \\
        l_2(e) & = \Vert e \Vert_2 = (\sum \limits _{i=1}^{n} e^2_{i})^{\frac{1}{2}} , \frac{\partial l_2(e) }{\partial e_i}  = k_1\cdot e_i, 
    \end{aligned}
    \right.
\label{eq:l1l2}
\end{equation}
where $e=\left |x-x'\right |$, denotes the distance between the reconstructed image and original image at a specific pixel during photometric consistency enforcement; $k_1 = (\sum \limits _{i=1}^{n} e^2_{i})^{-\frac{1}{2}}$, which can be considered as a constant for a specific $e_i$. 

\begin{figure}[t]
  \begin{center}        
     \includegraphics[trim={0cm 0.2cm 0cm 0.3cm},clip,width=1.0\linewidth]{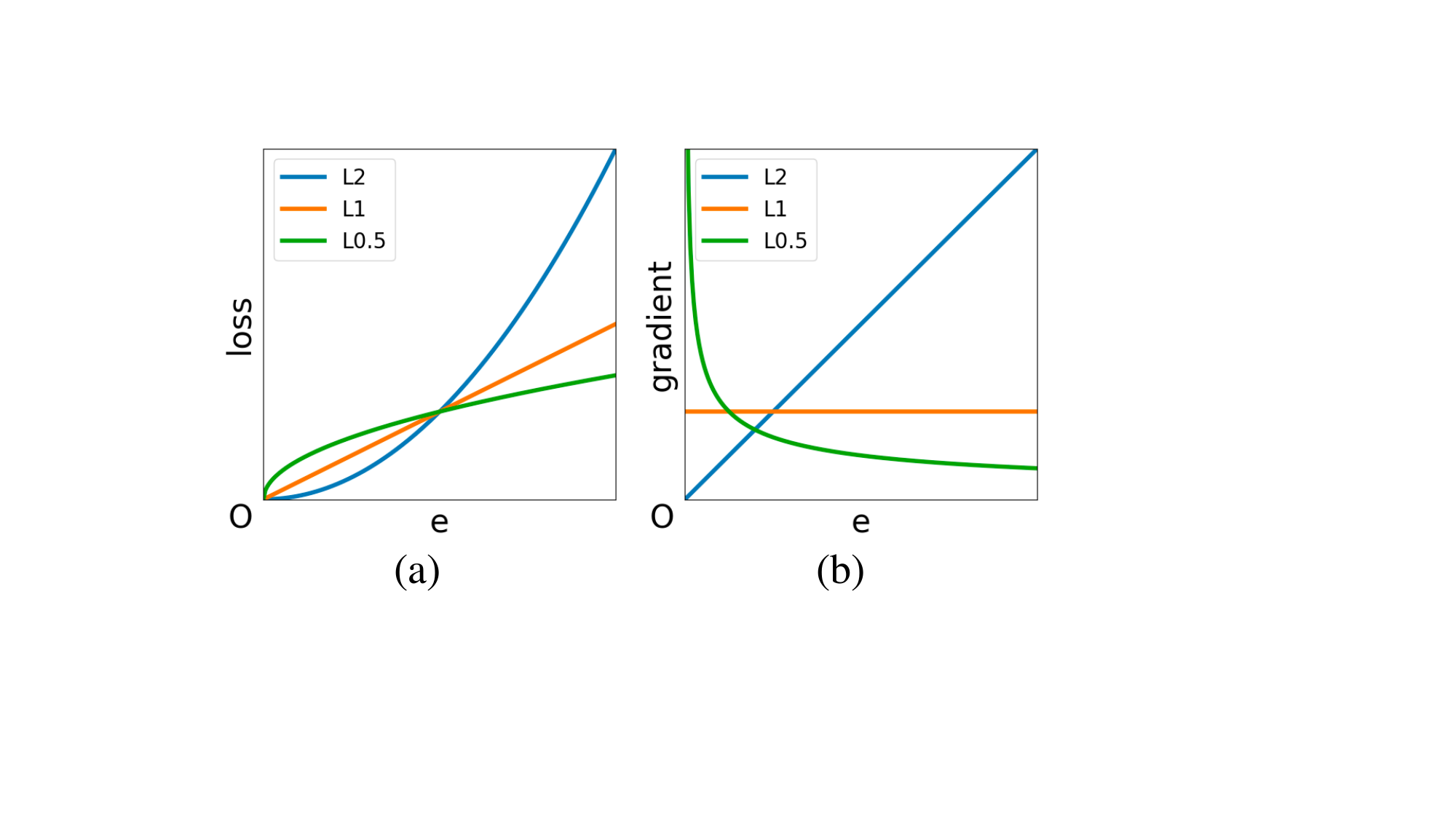}
  \end{center}
  \caption{\textbf{Comparison of different norms.} (a) Different norm-based losses about $e$. (b) The gradient of different norm-based losses with regard to $e$. The horizontal axis $e$ denotes the pixel-wise distance of image reconstruction with the estimated reference image depth. 
  The $\mathcal{L}2$ norm aims to reduce outliers, the $\mathcal{L}1$ norm treats all points equally, while the $\mathcal{L}0.5$ norm concentrates more on accurate points.}
  \label{fig:norms}
  \vspace{-0.3cm}
\end{figure}

During training, the parameters of the network are updated in a back-propagation manner, using gradients computed with respect to the loss function.
According to \cref{eq:l1l2}, we can draw a conclusion the $\mathcal{L}1$ norm is a fair norm that treats all points equally, aiming to obtain depth maps with low mean absolute error;  while the $\mathcal{L}2$ norm concentrates more on the pixels with larger error, to reduce the outliers, \ie, {\em terrible points}. 
However, these two norms cannot focus on {\em accurate points} directly.
Therefore, we propose a photometric consistency loss based on the $\mathcal{L}0.5$ norm. 
The $\mathcal{L}0.5$ norm and its gradient formula can be expressed as:
\begin{gather}
l_{0.5}(e) = \Vert e \Vert_{\frac{1}{2}} = (\sum \limits _{i=1}^{n} e_{i}^{\frac{1}{2}})^2, \frac{\partial l_{0.5}(e) }{\partial e_i} =k_2\cdot  e_i^{-\frac{1}{2}}, 
\end{gather}
in which $k_2 = \sum \limits _{i=1}^{n} e_i^{\frac{1}{2}}$.
And the comparison of different norms can be seen in \cref{fig:norms} more intuitively.

Unlike $\mathcal{L}1$-norm or $\mathcal{L}2$-norm, the $\mathcal{L}0.5$-norm pays more attention to {\em accurate points}.
Owing to its gradient property, applying this norm to the photometric consistency loss can make {\em accurate points} more accurate, turn {\em ordinary points} into {\em accurate points}, and pay less attention to {\em terrible points}.
According to the above analyses, we propose the $\mathcal{L}0.5$ photometric consistency loss as:
\begin{equation}
\resizebox{0.87\columnwidth}{!}{ $L_{0.5PC} \! = \! \sum \limits _{i=2}^{N} \! \frac
{\Vert (\hat{I}_i-I_1)\odot M_i \Vert_{\frac{1}{2}} \! +  \! \Vert(\nabla \hat{I}_i-\nabla I_1) \odot M_i \Vert_{\frac{1}{2}}}
{\Vert M_i \Vert_1}$. }
\end{equation}


\subsection{Overall Loss}
\label{sec:optimization}
The overall loss function of our proposed framework is constructed as follows:
\begin{align}
L =
& 
\lambda _1 L_{0.5PC} + \lambda _2 L_{ICC} + \lambda _3 L_{SCC} + 
\notag \\
&
\lambda _4 L_{SSIM} + \lambda _5 L_{Smooth}. 
\end{align}

$L_{SSIM}$ \cite{wang2004image} and $L_{Smooth}$ \cite{mahjourian2018unsupervised} denote structure similarity loss and depth smooth loss respectively, which are used broadly in previous unsupervised MVS methods.
The balancing weights are empirically set as $\lambda _1$ = 0.8, $\lambda _3$=0.01, $\lambda _4$= 0.2, $\lambda _5$= 0.0067, and the weight for image-level contrastive consistency is initialized with $\lambda _2$= 0.01 and doubles every two epochs. 
Note that the color fluctuation augmentation used by \cite{xu2021self,chang2022rc,xu2021digging} is also applied in the image-level contrastive branch.

\section{Experiment}
\label{sec:experment}

\subsection{Datasets}
\noindent \textbf{DTU Dataset  \cite{aanaes2016large}}  is a popular indoor benchmark comprising 79 training scans and 22 testing scans, all taken under 7 different lighting conditions.
The DTU dataset provides 3D point clouds captured by structured-light sensors, with each view consisting of an image and its calibrated camera parameters.
Following common practices,  we perform training on the provided training dataset, while evaluation is conducted on the designated evaluation dataset.


\paragraph{Tanks\&Temples \cite{knapitsch2017tanks}} 
is a large-scale dataset collected under realistic lighting conditions, consisting of an intermediate subset and an advanced subset. 
The intermediate subset includes 8 scenes, and the advanced subset includes 6 scenes. 
All scenes vary in terms of scale, surface reflection, and exposure conditions. 
These two subsets are widely used to verify the generalization performance of MVS methods.

\subsection{Implementation Details}
\paragraph{Training.} 
Our CL-MVSNet is trained on the DTU training set for 16 epochs in an end-to-end unsupervised learning manner. 
Following previous methods, the input image number $N$ is set to 5, and images are resized and cropped to 512 $\times$ 640.
Our backbone is similar to CasMVSNet with 3 stages, and the depth hypotheses for each stage are 48, 32, and 8 respectively.
We adopt Adam \cite{kingma2014adam} to optimize our model with an initial learning rate 0.0005, which is decayed by 2 after 10, 12, and 14 epochs. 
The network is implemented in Pytorch and trained on 8 NVIDIA Tesla V100s.


\begin{table}
  \caption{{\bf Depth map evaluation results in terms of accuracy on DTU evaluation set(higher is better).}  CL-MVSNet acquires the best depth estimation. All thresholds are given in millimeters.}
  \label{tb:depth_evaluation}
  \vspace{-0.2cm}
  \begin{center}
  \resizebox{0.65\linewidth}{!}{
  \begin{tabular}{lccc}
  \toprule
  Method & $\le$ 2 $\uparrow$ & $\le$ 4 $\uparrow$	& $\le$ 8 $\uparrow$ \\
  \midrule
  MVSNet \cite{Yao_2018_ECCV}  & 0.704 & 0.778	& 0.815 \\
  Unsup MVSNet \cite{khot2019learning} & 0.317 	& 0.384	& 0.402 \\
  M3VSNet \cite{huang2021m3vsnet} & 0.603 	& 0.769	& 0.857 \\
  JDACS-MS \cite{xu2021self} & 0.553 	& 0.705	& 0.786 \\
  RC-MVSNet \cite{chang2022rc} & 0.730 	& 0.795	& 0.863 \\
  \midrule
  \textbf{CL-MVSNet(Ours)} & \textbf{0.757} 	& \textbf{0.829}	& \textbf{0.868} \\
  \bottomrule
  \end{tabular}
  }
  \end{center}
  \vspace{-0.1cm}
\end{table}

\paragraph{Testing.} 
On the DTU testing set, the input image number $N$ is set to 5 as \cite{chang2022rc, ding2022kd}, and the resolution is 1184 $\times$ 1600 following \cite{yang2021self, yang2020cost,qiu2022self,Yao_2018_ECCV,gu2020cascade}.
The inferred depth maps are filtered with photometric and geometric consistencies and then fused to a point cloud following \cite{chang2022rc}.
On Tanks\&Temples, the input image number $N$ is set to 7, and the resolution is 1024 $\times$ 1920 or 768 $\times$ 576.
Note that our model trained on DTU is directly used to test on Tanks\&Temples without finetuning on BlendedMVS \cite{yao2020blendedmvs} as \cite{xu2021self,ding2022kd} or training on Tanks\&Temples training set as \cite{yang2021self}.
The inferred depth maps are filtered with photometric and geometric consistencies and then fused to a point cloud with the same strategy as \cite{chang2022rc}.
The number of depth hypotheses in the coarsest stage is set to 64, and the corresponding depth interval is set to 3 times as the interval of \cite{Yao_2018_ECCV}.
And only the regular branch works for testing.
More details can be found in the supplementary material.

\subsection{Benchmark Results on DTU}

\begin{table}
  \vspace{-0.3cm}
  \caption{{\bf Quantitative results on DTU evaluation set.} Best results in each category are in {\bf bold}. }
    \label{tb:dtu_compare}
  \vspace{-0.5cm}
  
  \begin{center}

    \renewcommand\arraystretch{1.1}
  \resizebox{1.0\linewidth}{!}{
  \begin{tabular}{clccc}
  \toprule
  & Method & Acc. $\downarrow$ & Comp.$\downarrow$ & Overall$\downarrow$ \\
  \midrule
  \multirow{6}*{Supervised}
  & SurfaceNet \cite{ji2017surfacenet} & 0.450 & 1.040 & 0.745 \\
  & MVSNet \cite{Yao_2018_ECCV} & 0.396 & 0.527 & 0.462 \\
  & CasMVSNet \cite{gu2020cascade} & 0.325 & 0.385 & 0.355 \\
  & PatchmacthNet \cite{wang2021patchmatchnet} & 0.427 & \textbf{0.277} & 0.352 \\
  & CVP-MVSNet \cite{yang2020cost} & {\bf 0.296} & 0.406 & 0.351 \\
  & UCS-Net \cite{cheng2020deep} & 0.338 & 0.349 & \textbf{0.344} \\
  \hline
  \multirow{3}*{\makecell{Multi-stage \\ Self-supervised}}
  & Self-sup CVP \cite{yang2021self} & {\bf 0.308} & 0.418 & 0.363 \\
  & U-MVSNet \cite{xu2021digging} & 0.354 & 0.3535 &  0.3537 \\
  & KD-MVS \cite{ding2022kd} & 0.359 & {\bf 0.295} & {\bf 0.327} \\
  \midrule
  \multirow{7}*{\makecell{End-to-end \\ Unsupervised}}
  & Unsup MVSNet \cite{khot2019learning} & 0.881 & 1.073 & 0.977 \\
  & MVS2 \cite{dai2019mvs2} & 0.760 & 0.515 & 0.637 \\
  & M3VSNet \cite{huang2021m3vsnet} & 0.636 & 0.531 & 0.583 \\
  & DS-MVSNet \cite{li2022ds} & \textbf{0.374} & 0.347 & 0.361 \\  
  & JDACS-MS \cite{xu2021self} & 0.398 & 0.318 & 0.358 \\
  & RC-MVSNet \cite{chang2022rc} & 0.396 & 0.295 & 0.345 \\
  & {\bf CL-MVSNet(Ours)} &  0.375 & {\bf 0.283} & {\bf 0.329} \\
  \bottomrule
  \end{tabular}
  }
  \end{center}
  \vspace{-0.6cm}

\end{table}

\begin{figure*}[t]
  \vspace{-0.8cm}
  \begin{center}
     \includegraphics[trim={0cm 0cm 0cm 0.1cm},clip,width=0.9\linewidth]{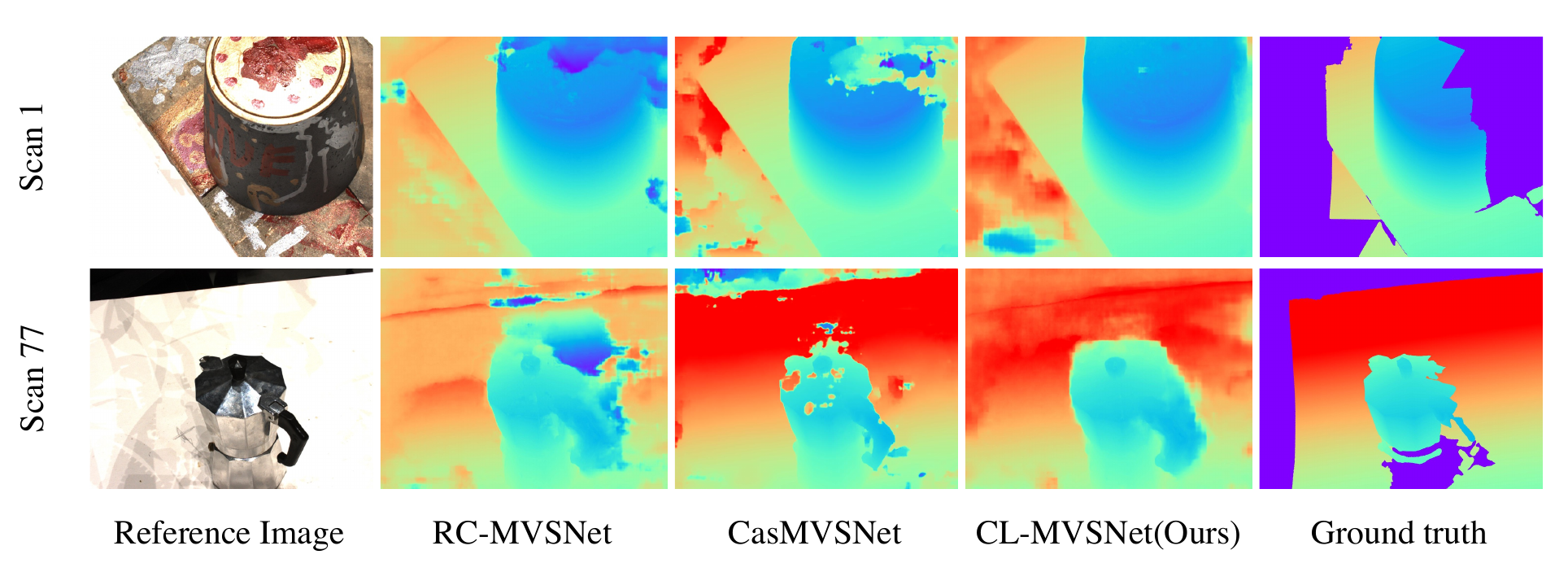}
  \end{center}
  \vspace{-0.4cm}
  \caption{{\bf Inferred depth map comparison on DTU \cite{aanaes2016large}.} Compared with the SOTA unsupervised method \cite{chang2022rc} and the supervised counterpart \cite{gu2020cascade}, CL-MVSNet gains more accurate depth maps.}
  \label{fig:depth_dtu}
\end{figure*}

\begin{figure*}[t]
  \vspace{-0.3cm}
  \begin{center}
     \includegraphics[trim={0cm 0cm 1cm 0cm},clip,width=1.0\linewidth]{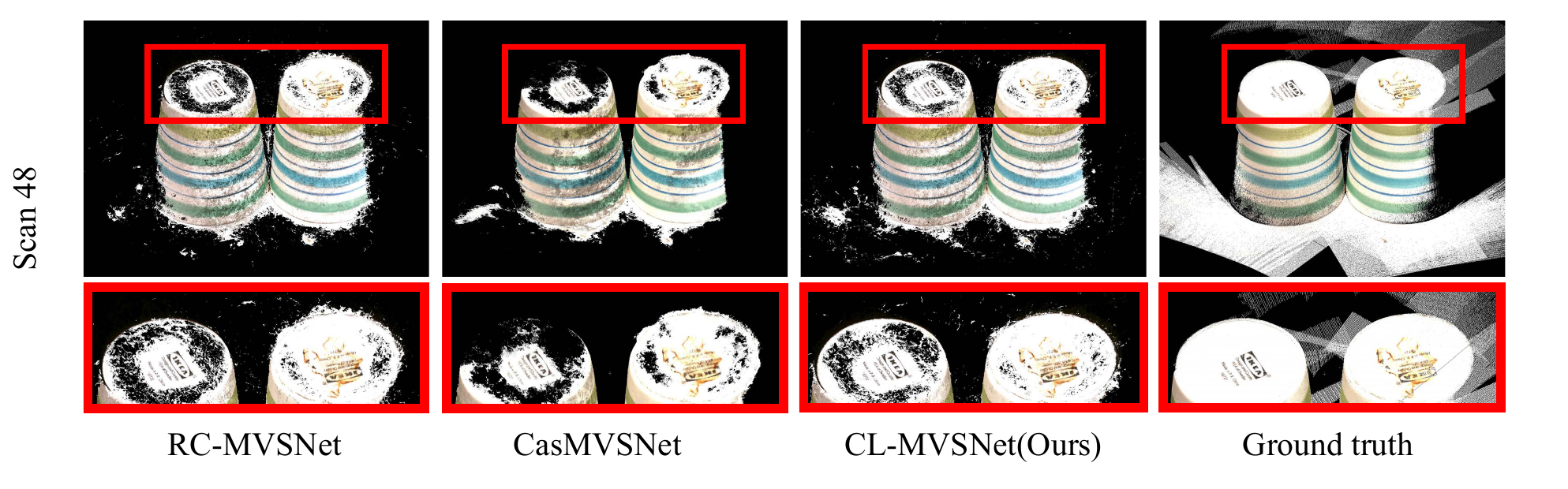}
  \end{center}
  \vspace{-0.5cm}
  \caption{\textbf{Comparison of reconstructed results on scan48 of DTU benchmark \cite{aanaes2016large}.} CL-MVSNet reconstructs more complete results than the SOTA unsupervised method \cite{chang2022rc} and the supervised counterpart \cite{gu2020cascade}.}
  \label{fig:comparison_dtu}
\end{figure*}

\begin{table*}[ht]

  \caption{{\bf Quantitative results of F-score on Tanks\&Temples benchmark.}}
  \label{tb:tank_compare}
  \vspace{-0.3cm}

  \begin{center}
  \renewcommand\arraystretch{1.1}
  \resizebox{1.0\linewidth}{!}{
  \begin{tabular}{clccccccccc|ccccccc}
  \toprule
  &  & \multicolumn{9}{c|}{{Intermediate }} & \multicolumn{7}{c}{Advance }\\
  \midrule
  & Method & Mean$\uparrow$ & Fam.$\uparrow$ & Fra.$\uparrow$ & Hor.$\uparrow$ & Lig.$\uparrow$ & M60$\uparrow$ & Pan.$\uparrow$ & Pla.$\uparrow$ & Tra.$\uparrow$ & Mean$\uparrow$ & Aud.$\uparrow$	& Bal.$\uparrow$	& Cou.$\uparrow$	& Mus.$\uparrow$	& Pal.$\uparrow$	& Tem.$\uparrow$ \\

  \midrule
  \multirow{4}*{{Traditional}} 
  & COLMAP \cite{schonberger2016structure,schonberger2016pixelwise} & 42.14 & 50.41 & 22.25 & 25.63 & 56.43 & 44.83 & 46.97 & 48.53 & 42.04 & 27.24	& 16.02 &	25.23 &	34.70 &	41.51 &	18.05 &	27.94 \\
  & ACMM \cite{xu2019multi}& 57.27 & 69.24 & 51.45 & 46.97 & 63.20 & 55.07 & 57.64 & \textbf{60.08} & 54.48 & 34.02	&	23.41	& 32.91	& 41.17	& 48.13 &	23.87	& 34.60\\
  & ACMP \cite{xu2020planar} & 58.41 & 70.30 & 54.06 & \textbf{54.11}	& 61.65	& 54.16	& 57.60	& 58.12	& 57.25 & 
  37.44	&	\textbf{30.12} &	34.68 &	\textbf{44.58} &	50.64 &	27.20 & 37.43\\
  & ACMMP \cite{xu2022multi}& \textbf{59.38} & \textbf{70.93}	& \textbf{55.39}	& 51.80 &	\textbf{63.83} &	\textbf{55.94} &	\textbf{59.47}	& 59.51	& \textbf{58.20} & \textbf{37.84}	&	30.05 &	\textbf{35.36} &	44.51 &	\textbf{50.95}	& \textbf{27.43} &	\textbf{38.73} \\

  \midrule
  \multirow{5}*{{Supervised}} 
  & MVSNet \cite{Yao_2018_ECCV} & 43.48 & 55.99 & 28.55 & 25.07 & 50.79 & 53.96 & 50.86 & 47.90 & 34.69 & -& -& -& -& -&- & -\\

  & PatchmatchNet \cite{wang2021patchmatchnet} & 53.15 & 66.99 & 52.64 & 43.24 & 54.87 & 52.87 & 49.54 & 54.21 & \textbf{50.81} & \textbf{32.31} & \textbf{23.69}	& 37.73	& \textbf{30.04}	& 41.80	& \textbf{28.31}	& \textbf{32.29 }\\
  & CVP-MVSNet \cite{yang2020cost} & 54.03 & \textbf{76.50} & 47.74 & 36.34 & 55.12 & \textbf{57.28} & \textbf{54.28} & 57.43 & 47.54 & -& -& -& -& -&- & -\\
  & UCS-Net \cite{cheng2020deep} & 54.83 & 76.09 & 53.16 & 43.03 & 54.00 & 55.60 & 51.49 & 57.38 & 47.89 & -& -& -& -& -&- & -\\
  & CasMVSNet \cite{gu2020cascade} & \textbf{56.42} & 76.36 & \textbf{58.45} & \textbf{46.20} & \textbf{55.53} & 56.11 & 54.02 & \textbf{58.17} & 46.56 & 31.12 & 19.81	& \textbf{38.46}	& 29.10	& \textbf{43.87}	& 27.36	& 28.11\\
  \midrule
  \multirow{3}*{{\makecell{Multi-stage \\ Self-supervised}}}
  & Self-sup CVP \cite{yang2021self} & 46.71 & 64.95 & 38.79 & 24.98 & 49.73 & 52.57 & 51.53 & 50.66 & 40.45 & -& -& -& -& -&- & -\\
  & U-MVSNet \cite{xu2021digging} & 57.15 & 76.49 & 60.04 & 49.20 & 55.52 & 55.33 & 51.22 & 56.77 & 52.63 & -& -& -& -& -&- & -\\
  & KD-MVS \cite{ding2022kd} & \textbf{64.14} & \textbf{80.42}	& \textbf{67.42}	& \textbf{54.02}	& \textbf{64.52}	& \textbf{64.18}	& \textbf{61.60}	& \textbf{62.37}	& \textbf{58.59} & \textbf{37.96} & \textbf{27.24}	& \textbf{44.10}	& \textbf{35.47}	& \textbf{49.16}	& \textbf{34.68}	& \textbf{37.11} \\
  \midrule
  \multirow{6}*{{\makecell{End-to-end \\ Unsupervised}}}
  & MVS2 \cite{dai2019mvs2} & 37.21 & 47.74 & 21.55 & 19.50 & 44.54 & 44.86 & 46.32
& 43.38 & 29.72 & -& -& -& -& -&- & -\\
  & M3VSNet \cite{huang2021m3vsnet} & 37.67 & 47.74 & 24.38 & 18.74 & 44.42 & 43.45 & 44.95
& 47.39 & 30.31 & -& -& -& -& -&- & -\\
  & JDACS-MS \cite{xu2021self} & 45.48 & 66.62 & 38.25 & 36.11 & 46.12 & 46.66 & 45.25
& 47.69 & 37.16 & -& -& -& -& -&- & - \\

  & DS-MVSNet \cite{li2022ds} & 54.76 & 74.99 & 59.78 & 42.15 & 53.66 & 53.52 & 52.57 & 55.38 & 46.03 & -& -& -& -& -&- & - \\

  & RC-MVSNet \cite{chang2022rc} & 55.04 & 75.26 & 53.50 & 45.52 & 53.49 & 54.85 & 52.30 & 56.06 & 49.37 & 30.82 & 21.72	& 37.22	& 28.62	& 37.37	& 27.88	& \textbf{32.09} \\
  & {\bf CL-MVSNet(Ours)} & \textbf{59.39} & \textbf{76.35}	& \textbf{62.37}	& \textbf{49.93}	& \textbf{60.02}	& \textbf{57.44}	& \textbf{59.97}	& \textbf{56.74}	& \textbf{52.28} & \textbf{37.03} & \textbf{28.07}	& \textbf{43.55}	& \textbf{37.47}	& \textbf{50.86}	& \textbf{31.45}	& 30.78\\
  \bottomrule
  \end{tabular}
  }
  \end{center}
\end{table*}

\begin{figure*}[t]
  \vspace{-0.6cm}
  \begin{center}
     \includegraphics[trim={0cm 0cm 0cm 3mm},clip,width=0.9\linewidth]{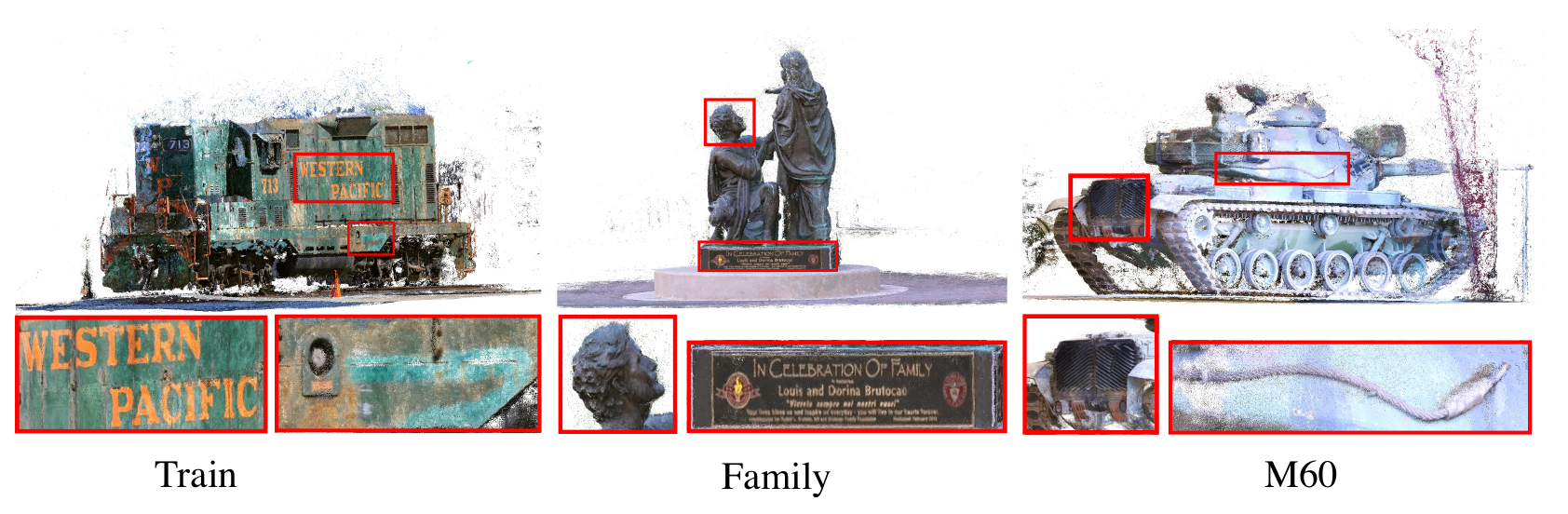}
  \end{center}
  \vspace{-0.5cm}
  \caption{\textbf{Qualitative results on Tanks\&Temples \cite{knapitsch2017tanks}.}  }
  \vspace{-0.1cm}
  \label{fig:tank_point}
\end{figure*}

\begin{figure}
  \vspace{-0.3cm}
  \begin{center}
  
     \includegraphics[trim={0cm 0cm 0cm 0cm},clip,width=1.0\linewidth]{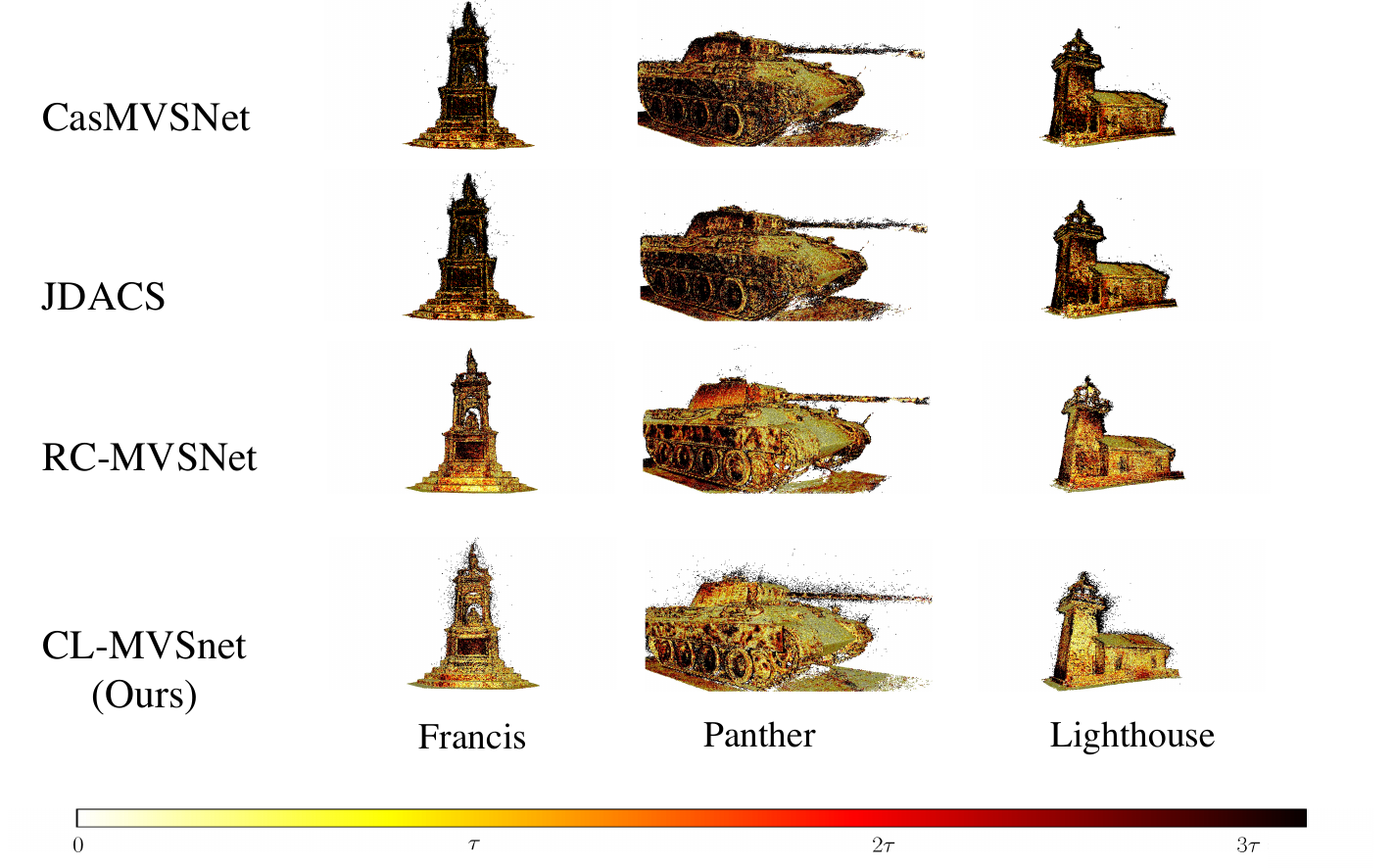}
     
  \end{center}
  \vspace{-0.3cm}
  \caption{\textbf{
  Qualitative comparison of reconstruction quality on Tanks\&Temples benchmark  \cite{knapitsch2017tanks}.} 
   Darker regions contain larger errors. 
  CL-MVSNet yields better performance than the SOTA unsupervised methods \cite{xu2021self,chang2022rc} and the supervised counterpart \cite{gu2020cascade}.
  }
  \label{fig:comparison_on_tnt}
  \vspace{-0cm}
\end{figure}

We compare the depth map evaluation results on the DTU evaluation set as shown in \cref{tb:depth_evaluation}.
Then we compare the quantitative reconstruction results on DTU as shown in \cref{tb:dtu_compare}, our CL-MVSNet architecture achieves the best completeness and overall score among all end-to-end unsupervised methods.
It is worth noting that CL-MVSNet also outperforms its supervised counterpart CasMVSNet \cite{gu2020cascade}.
The qualitative results of the depth estimation shown in \cref{fig:depth_dtu} and the 3D point cloud reconstruction shown in \cref{fig:comparison_dtu} also verify the advantages of our model.

\subsection{Benchmark Results on Tanks\&Temples}

In line with existing methods, we test our method on the Tanks\&Temples benchmark to verify the generalization ability. 
The quantitative results on Tanks\&Temples are reported in \cref{tb:tank_compare}. 
Our method achieves SOTA performance among all existing end-to-end unsupervised MVS methods. 
Our method also surpasses the supervised counterpart CasMVSNet \cite{gu2020cascade}. 
The qualitative reconstruction results are visualized in \cref{fig:tank_point}.
\cref{fig:comparison_on_tnt} shows a qualitative comparison of reconstruction quality with other methods. The performance on Tanks\&Temples shows the generalizability and robustness of our model.

\subsection{Ablation Study}
\label{sec:ablation}

We perform an ablation study to confirm the effectiveness of each part in our model under different configurations as shown in \cref{tb:ablation}. 
Applying the image-level contrastive consistency loss to the model leads the model to be more context-aware for indistinguishable regions, which enhances the robustness and generalizability. 
Besides, the model combined with the scene-level contrastive consistency loss tends to perform better in areas of view-dependent effects, thus gaining more accurate and complete reconstruction results.
Moreover, $\mathcal{L}0.5$ photometric consistency loss further advances the CL-MVSNet, which guides to model to focus on accurate points, resulting in an improvement in the accuracy of the model.
More ablation studies are included in Supplementary Materials.

\subsection{Efficiency Comparison to SOTA Multi-stage Self-supervised Method}
As aforementioned, multi-stage self-supervised methods cannot be trained end-to-end.
Specifically, KD-MVS \cite{ding2022kd} takes several training rounds for distillation, needs complex pre-training and pre-processing before each round, uses additional dataset \cite{yao2020blendedmvs} for pseudo label generation, and requires extra storage space to save the pseudo labels.
Without the above limitations, our method adopts the same backbone as theirs, but converges and achieves competitive results in just 16 epochs, with a training time of only 10 hours per epoch on one single NVIDIA Tesla V100.

\begin{table}
  \vspace{-0.1cm}
  \caption{{\bf Ablation study of different components of our proposed CL-MVSNet on DTU \cite{aanaes2016large}.} }
  \label{tb:ablation}
  \vspace{-0.5cm}
      \renewcommand\arraystretch{1.1}
  \begin{center}
  \resizebox{1.0\linewidth}{!}{
  \begin{tabular}{ccccc|ccc}
  \toprule
  $L_{PC}$ & $L_{DA}$ \cite{xu2021self}  & $L_{ICC}$ & $L_{SCC}$ & $L_{0.5PC}$  & Acc.$\downarrow$ & Comp.$\downarrow$ & Overall$\downarrow$ \\
  \midrule
  \midrule
  \checkmark  & \checkmark  &   &  &   & 0.422 & 0.334  & 0.378 \\
  \checkmark  &   & \checkmark  &  &   & 0.403 & 0.315  & 0.359 \\
  \checkmark  &   & \checkmark  & \checkmark &  & 0.392 & 0.286  & 0.339 \\
    &   & \checkmark  & \checkmark & \checkmark  & \textbf{0.375} & \textbf{0.283}  & \textbf{0.329} \\
  \bottomrule
  \end{tabular}
  }
  \end{center}
  \vspace{-0.9cm}
\end{table}
  \vspace{-0.2cm}

\section{Conclusion and Limitation}
\label{sec:conclusions}


\noindent {\bf Conclusion.} We have presented an effective unsupervised approach for Multi-View Stereo, termed as CL-MVSNet, which leverages dual-level contrastive learning to handle the issues of indistinguishable regions and view-dependent effects. 
For indistinguishable regions, we propose an image-level contrastive branch to encourage the model to take more contextual information into account.
For view-dependent effects, a scene-level contrastive branch is adopted to boost the robustness. 
Besides, we explore an $\mathcal{L}0.5$ photometric consistency loss to emphasize the penalty of accurate points, resulting in more accurate reconstruction. 
We experimentally demonstrate that CL-MVSNet outperforms all SOTA end-to-end unsupervised MVS methods and the supervised counterpart on the DTU \cite{aanaes2016large} and Tanks\&Temples \cite{knapitsch2017tanks} benchmarks.

\noindent {\bf Limitation.} 
Our model has addressed the limitations of indistinguishable regions and view-dependent effects, but the accurate depth estimation in object edge areas remains a challenge. 
It is worth noting that this is a common problem in unsupervised MVS methods.
To mitigate this issue, we adopt an edge-aware depth smoothness loss proposed in \cite{khot2019learning}, which is based on the assumption that the gradient maps of the input reference image and the inferred depth map should be similar. 
However, this simple assumption may be invalid in many cases. 
For instance, there may be significant color gradient changes within the same object.

\noindent {\bf Acknowledgements.} This work is financially supported by National Natural Science Foundation of China U21B2012 and  62072013, Shenzhen Science and Technology Program-Shenzhen Cultivation of Excellent Scientific and Technological Innovation Talents project(Grant No. RCJC20200714114435057), Shenzhen Science and Technology Program-Shenzhen Hong Kong joint funding project (Grant No. SGDX20211123144400001), this work is also financially supported for Outstanding Talents Training Fund in Shenzhen. 


{\small

}

\end{document}